\documentclass{article}

\usepackage[preprint,nonatbib]{neurips_2022}

\usepackage[utf8]{inputenc} 
\usepackage[T1]{fontenc}    
\usepackage{hyperref}       
\usepackage{url}            
\usepackage{booktabs}       
\usepackage{amsfonts}       
\usepackage{nicefrac}       
\usepackage{microtype}      
\usepackage[dvipsnames]{xcolor}
\usepackage{soul}
\usepackage{algorithm}
\usepackage{algpseudocode}
\usepackage{algorithmicx}
\usepackage{mathtools}
\usepackage{amsmath}	
\usepackage[thinc]{esdiff}
\usepackage{xspace}
\usepackage{booktabs}

\newcommand{\aname}{{\small ILD}\xspace}
\newcommand{\methodname}{{ Imitation Learning via Differentiable Physics}\xspace}

\renewcommand{\eqref}[1]{(\ref{#1})}
\newcommand{\figref}[1]{Fig.~\ref{#1}}

\newcommand{\tabref}[1]{Table~\ref{#1}}

\title{\methodname}

\author{%
  Siwei Chen$^{12}$, Xiao Ma$^{1}$, Zhongwen Xu$^{1}$\\
  $^1$Sea AI Lab $^2$National University of Singapore \\
  \texttt{siwei-15@comp.nus.edu.sg} \\
  \texttt{\{max,xuzw\}@sea.com} \\
}

\begin{document}

\maketitle

\begin{abstract}

Existing imitation learning (IL) methods such as inverse reinforcement learning (IRL) usually have a double-loop training process, alternating between learning a reward function and a policy and tend to suffer long training time and high variance. 
In this work, we identify the benefits of differentiable physics simulators and propose a new IL method, i.e., \methodname (\aname), which gets rid of the double-loop design and achieves significant improvements in final performance, convergence speed,  and stability.
The proposed \aname incorporates the differentiable physics simulator as a \textit{physics prior} into its computational graph for policy learning.
It unrolls the dynamics by sampling actions from a parameterized policy, simply minimizing the distance between the expert trajectory and the agent trajectory, and back-propagating the gradient into the policy via temporal physics operators. With the physics prior, \aname policies can not only be transferable to unseen environment specifications but also yield higher final performance on a variety of tasks. In addition, \aname naturally forms a single-loop structure, which significantly improves the stability and training speed.
To simplify the complex optimization landscape induced by temporal physics operations, \aname dynamically selects the learning objectives for each state during optimization.
In our experiments, we show that \aname outperforms state-of-the-art methods in a variety of continuous control tasks with Brax, requiring only one expert demonstration. In addition, \aname can be applied to challenging deformable object manipulation tasks and can be generalized to unseen configurations. \footnote{The code can be found here: \url{https://github.com/sail-sg/ILD}}

\end{abstract}

\section{Introduction}


In a variety of applications ranging from games to real-world robotic tasks~\cite{ho2016generative, fu2017learning, zeng2020transporter}, imitation learning (IL) is popularly applied. However, collecting high-quality expert data is expensive, and existing IL methods tend to suffer long training time, unstable training process, high variance of learned IL policies, and suboptimal final performance.




Classical behavioral cloning (BC) methods learn policies directly from labeled data, but often suffer the covariate shift problem. 
This problem can be tackled in DAGGER~\cite{ross2011reduction} by interacting with the environment and querying experts online, which however requires significant human effort to label the actions. 
Other IL methods mainly include inverse reinforcement learning (IRL), adversarial imitation learning (AIL), and combinations of them.
IRL learns a reward function to match expert demonstrations~\cite{ziebart2008maximum,dadashi2020primal,hoshino2021opirl}, and AIL learns a discriminator to identify whether the action comes from an expert demonstration \cite{ho2016generative,kostrikov2018discriminator}. 
However, both IRL and AIL learn an additional intermediate signal, which introduces three main limitations: 1) the intermediate signal learning leads to a double-loop training process, which means 
long training time and complex implementation; 
2) the learning signal is a noisy and frequently updated moving target, and as a result, the policy learning tends to have a high variance; 
3) the intermediate signal, e.g., the reward function in IRL, inevitably loses the rich information embedded in the trajectories, e.g., environment dynamics.

In this work, we propose a new approach to IL, named \methodname (\aname), which recovers expert behavior by exploiting the Differentiable Physics Simulator (DPS)~\cite{freeman2021brax,hu2019taichi}.
Different from standard environments, DPS implements low-level physics operations with a differentiable function and allows the gradients to flow through the dynamics.
\aname takes advantage of DPSs by considering the environment dynamics as a \textit{physics prior} and incorporating it into its computational graph during back-propagation of the policy, such that the learned policy fully captures both the expert demonstration and the environment specifications. 
To achieve this, \aname simply minimizes the state-wise distance of a rollout trajectory generated by a parameterized policy to the expert demonstration, which also gives a single-loop design and avoids learning intermediate signals.
Nevertheless, the gradients of physics operators are highly non-convex, which often introduces a complex optimization landscape, and consequently, a naive implementation is often stuck in local minimum~\cite{freeman2021brax}.
To alleviate this issue, we introduce a simple yet effective Chamfer-$\alpha$ distance for trajectory matching.
For each state in the rollout trajectory, instead of exactly matching the corresponding expert state, we dynamically select the easiest local goal as the optimization target and gradually proceed to the harder ones as training progresses.
Chamfer-$\alpha$ distance naturally forms a curriculum learning setup, simplifies the optimization task, and eventually gives better final performance.



\begin{table}[]
\centering
\caption{\small Useful Properties among IL Methods}
\fontsize{9}{9}\selectfont
\begin{tabular}{lccc}
\toprule

 Property / Method Family                                     & IRL         & AIL         & \aname (ours)                          \\ \midrule
Layers of training loop              & Double-loop & Double-loop & Single-loop                   \\
Source of the learning signal     & Reward function           & Discriminator           & Differentiable dynamics \\
Transferability in changing dynamics & Depends     & No          & Yes                           \\ \bottomrule

\label{table:property}
\end{tabular}
\end{table}


A short comparison of some useful properties of the IL methods can be found in \tabref{table:property}. In contrast to the IRL and AIL methods, \aname does not introduce new intermediate signals and therefore requires no switching between policy learning and intermediate signal learning.
In terms of the learning paradigm, IRL learns a reward function, AIL learns a discriminator, and \aname uses the differentiable dynamics which makes the learned policy aware of the environment dynamics and transferable to unseen environment configurations.

Empirically, we validate \aname on a set of MuJoCo-like continuous control tasks from Brax~\cite{freeman2021brax} and a challenging cloth manipulation task. 
We show that \aname achieves significant improvements over the state-of-the-art IRL and AIL methods in terms of convergence time, training stability, and final performance. 
Given a fixed one-hour training time, \aname achieves 36\% higher performance based on the normalized score over all the tasks and baselines.

\section{Related Works}

\textbf{Imitation Learning}. 
Classical imitation learning methods directly imitate the expert demonstrations via Behaviour Cloning \cite{pomerleau1991efficient, ross2010efficient}. However, they often suffer covariant shift problems due to insufficient expert training data. 
The modern approach GAIL~\cite{ho2016generative} uses the idea of generative adversarial networks to learn a discriminator that distinguishes between learner trajectories and expert trajectories. 
The agent explores the environment and learns to mimic the expert's trajectory. SAM~\cite{sasaki2018sample} and DAC~\cite{kostrikov2018discriminator} continue the idea of GAIL and address the sample efficiency problem. OPOLO~\cite{zhu2020off} also proposes a sample efficient learning from the observation (LfO) approach that allows for non-policy-based optimization. 

\textbf{Inverse Reinforcement Learning (IRL)}. IRL is a type of imitation learning that learns policies by recovering reward functions to match the trajectories demonstrated by experts \cite{argall2009survey}. 
Early IRL methods such as MaxEntIRL \cite{boularias2011relative,ziebart2008maximum} minimize the KL divergence between the learner trajectory distribution and the expert trajectory distribution in the maximum entropy RL framework.
However, those IRL methods often involve a double-loop learning process in which the outer loop learns the reward function and the inner loop solves the forward RL learning problem. 
AIRL \cite{fu2017learning} builds on the adversarial learning idea of GAIL by learning a reward function for reinforcement learning. Moreover, OPIRL~\cite{hoshino2021opirl} learns an off-policy reward function and solves the sample inefficiency problem. Recently, PWIL~\cite{dadashi2020primal} proposes to learn the reward function by measuring the Wasserstein distance between the learner and the expert, achieving state-of-the-art results. 

\textbf{Differentiable Dynamics for Policy Learning}.
The differentiability of dynamics models has been explored to improve the stability and sample efficiency of policy learning. A commonly used paradigm is to learn a parameterized generative dynamics model by reconstructing the trajectory observations and train a policy by ``imagined'' trajectories with the learned generative model~\cite{hafner2019dream,ma2020contrastive,hafner2020mastering,clavera2020model}. However, built upon a learned model, they suffer temporal accumulative model error and long training time by switching over two loops for model learning and policy learning~\cite{yu2020mopo}. Recent advances in differentiable physics have shown their potential for policy learning by back-propagating the gradients through the physics operators~\cite{hu2019taichi,lin2022diffskill,huang2021plasticinelab,freeman2021brax}.
Different from a learned model, differentiable physics provides a ground-truth understanding of the environment dynamics and naturally guarantees a good generalization.  Nevertheless, the back-propagation through long temporal non-convex physics operators introduces a complex optimization landscape for policy learning, and as a result, a learned policy tends to be stuck at local minimums~\cite{freeman2021brax}. 

In contrast to existing methods, \aname avoids learning intermediate signals by computing the analytical learning gradient directly from the expert demonstration through differentiable physics. The analytical gradients carry rich information about both the expert intentions, i.e., the reward, and the specifications of the environment dynamics. Meanwhile, \aname dynamically selects local optimization goals for each state in the rollout trajectory, which gives a simpler optimization landscape for policy learning.

\section{Imitation Learning via Differentiable Physics}
\label{sec:method}

\begin{figure}[t]
	\centering
	\begin{tabular}{c c}
    \includegraphics[width=0.45\linewidth]{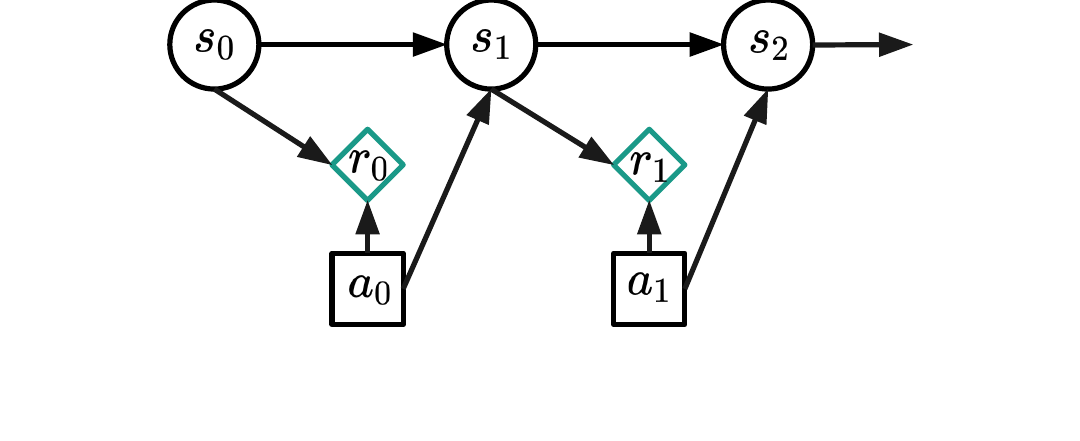} &
    \includegraphics[width=0.48\linewidth]{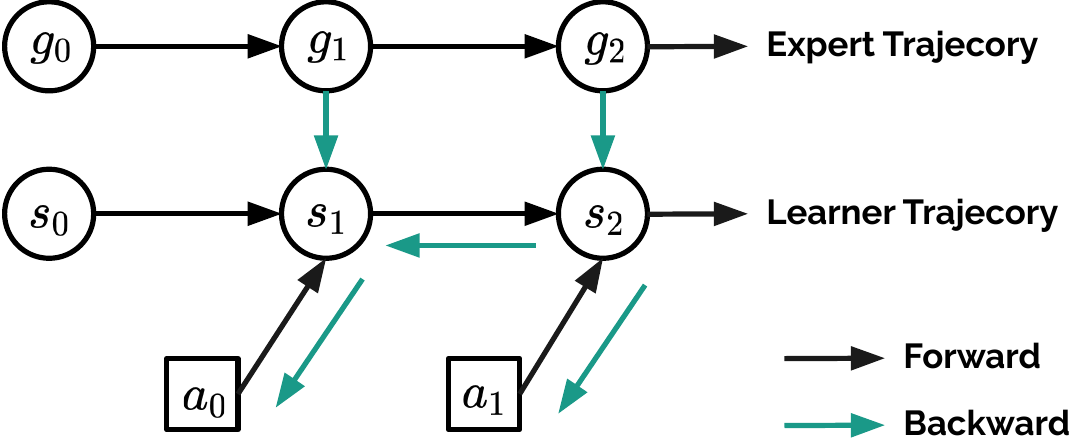} \\
    & \\
    (a) \small MDP Computation Graph & (b) \small ILD Computation Graph
    
    
	\end{tabular}
	\centering
	\caption{
	\small Computational graph of MDP and \aname. A typical Markov decision process (MDP) includes a reward function to evaluate the performance of a policy and provide learning signals to the learner agent. In our approach, we eschew the reward function and use differentiable dynamics to dynamically match expert states and make gradients flow back into the action. This design provides two main benefits: 1) we move away from the double-loop design in IRL and AIL, avoiding the process of alternating between learning rewards and learning policies; 2) the analytic gradient from the dynamics provides richer information than a single reward number, which guides the improvement of actions in local regions, bringing less variance in training and better performance.}
	\label{fig:computation}
\end{figure}



We propose \methodname (\aname), which learns from expert demonstrations via differentiable physics without any additional intermediate signals, e.g., reward functions in IRL. 
We assume that the underlying transition function of the task is built on a set of physics rules so that the cumulative compound error is small compared to a dynamics model learned from data. 
Take a point-mass system as an example:
\begin{equation*}
    x_{t+1} = x_t + \Delta t \cdot v_t \;\;\;\;\;  v_{t+1} = v_t + \Delta t \cdot  \frac{f}{m}
\end{equation*}
where the force $f$ is the action input to the system, $v_t$ is the speed and $x_t$ is the position of the point with mass $m$. The new position $x_{t+1}$ can be computed analytically based on the physics properties such as the mass $m$. 
More importantly, such a system is differentiable and can carry the gradient from the output state $x_{t+1}$ directly to the input force $f$.
The same idea generalizes to more complex physics systems such as dynamics of deformable objects like cloth and liquid. For simplicity, we abuse the notation of physics and dynamics in this work. The gist of \aname is to consider the differentiable dynamics as a \textit{physics prior} and incorporate it into the computational graph for policy learning.
With differentiable dynamics, future states in the trajectory can exert influence on early actions. In this way, \aname can leverage the rich information from future states and learn a policy that is aware of the environment specifications.
However, having a rich set of information does not necessarily lead to an optimal policy due to the complex optimization landscape through BPTT~\cite{freeman2021brax}. Therefore, we further decompose the optimization problem into many small and simple sub-steps by selecting suitable local learning such that each local goal can be effectively learned via differentiable dynamics.



\begin{figure}[t]
	\centering
	\begin{tabular}{c c c}
    \includegraphics[width=0.3\linewidth]{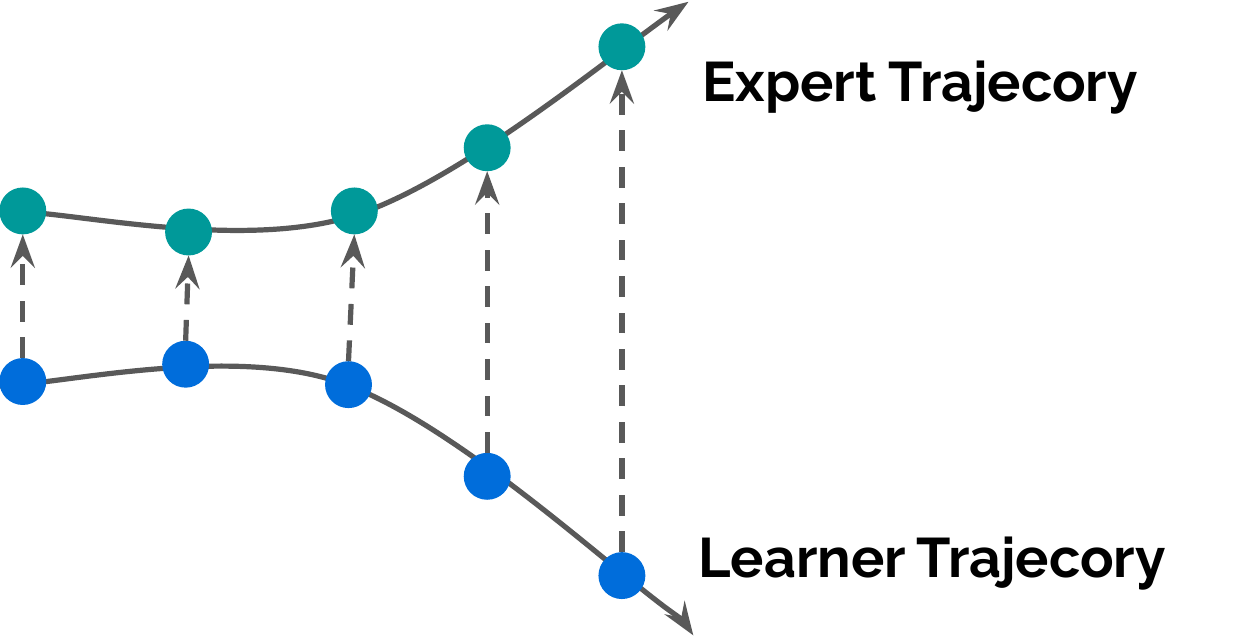} &
    \includegraphics[width=0.3\linewidth]{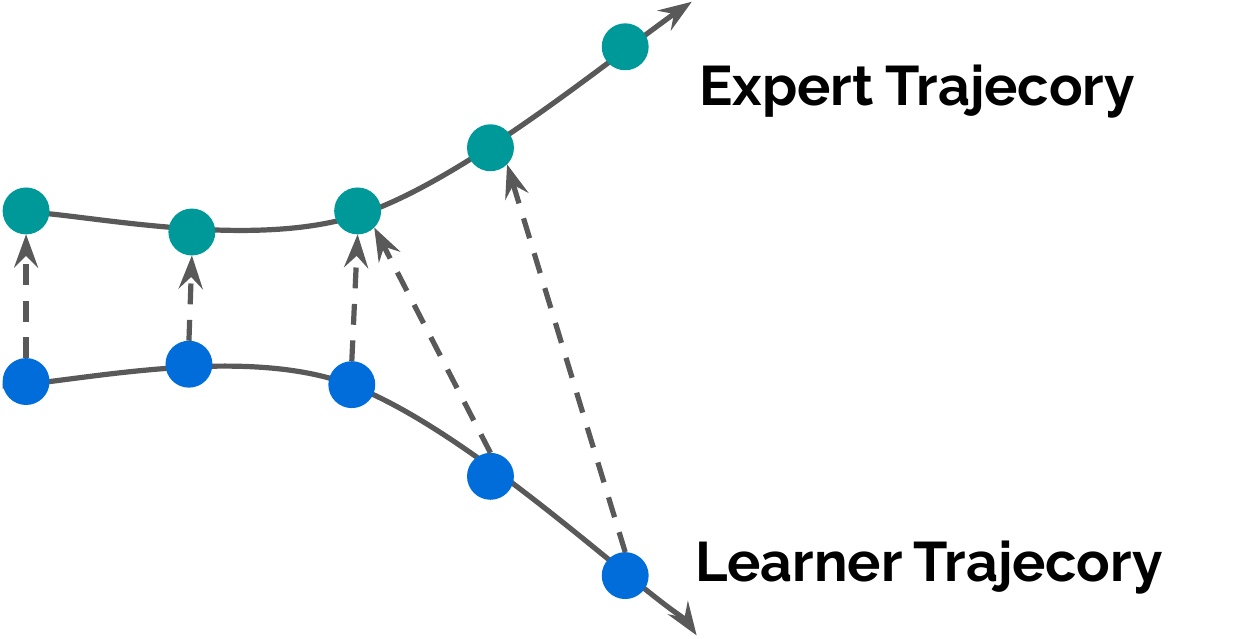} & 
    \includegraphics[width=0.3\linewidth]{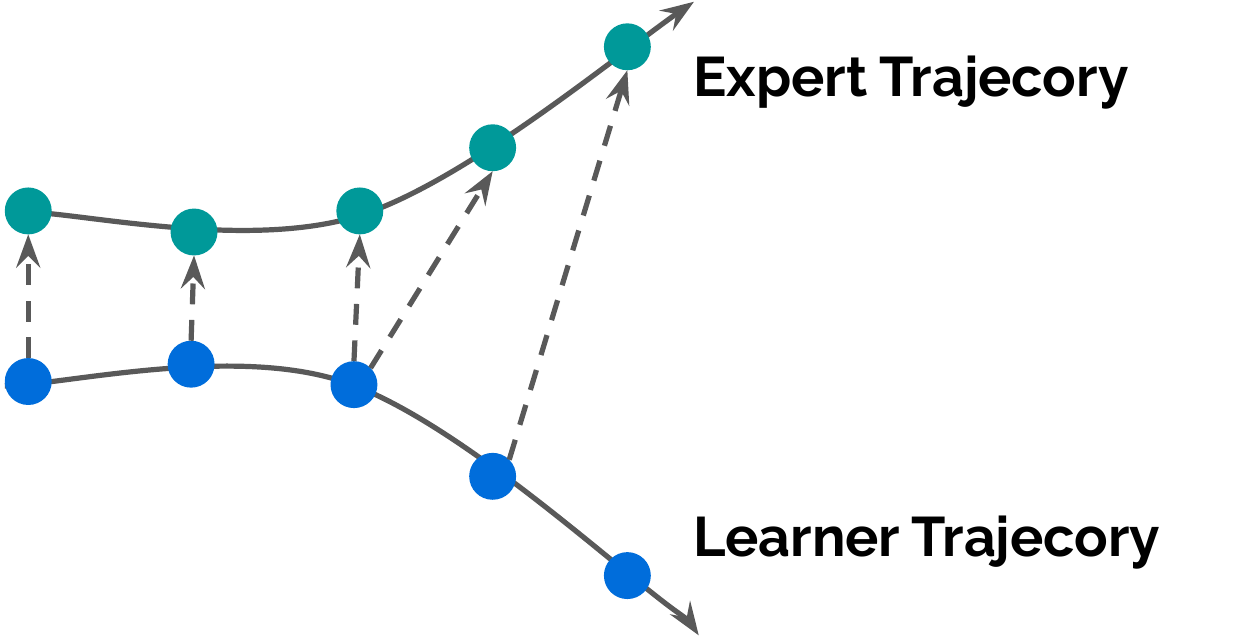} \\
    \\
    (a) \small L2 Loss & (b) \small Deviation Loss & (c) \small Coverage Loss
    
	\end{tabular}
	\centering
	\caption{\small Illustration of different loss functions. Both green dots and blue dots are the states in their trajectories. The L2 loss has a one-to-one matching, but the learning goal can be extremely difficult when later states deviate too much from the expert trajectory. The deviation loss only matches the closest state in the expert trajectory to reduce the difficulty of the learning goals, constraining the exploration space to be close to the expert trajectory. The coverage loss pulls the nearest states in the learner trajectory to be close to every state in the expert trajectory.  }
	\label{fig:loss_demo}
\end{figure}

\subsection{Differentiable Physics as Computational Graphs}
The computation graph of our method can be found in \figref{fig:computation}. In a detailed view, our method \aname first rolls out the dynamics with the learner agent $\pi_{\theta}$ to interact with the environment to collect state trajectories. At each step $t$ the learner policy  $\pi_\theta(a_t | s_t)$ outputs the means and variances for all the action dimensions to sample actions, and the reparameterization trick \cite{kingma2013auto} is used to allow the gradient to flow through the sampling process.
By iteratively unrolling the dynamics with the learner policy $\pi_\theta$, we observe a trajectory $\tau_\theta$ in the form of a list of states $s_{0:H}$ and a list of actions $a_{0:H}$. Treating the environment dynamics as a function $T(s_{t+1} | s_t, a_t)$, we can view the trajectory unrolling process as a temporal computation graph, $s_{1:H} = G(s_0, a_{0:H})$, where $s_0$ and $a_{0:H}$ are the inputs and $s_{1:H}$ denotes the outputs. Therefore, we can compute gradients regarding the action inputs. Since actions $a_t$ are conditioned on the policy parameters $\theta$, the entire computation graph $G$ can be reduced to $s_{1:H} = G(s_0, \theta)$ that outputs a list of states $s_{1:H}$ and a list of actions $a_{0:H}$. Most importantly, by the virtue of differentiable dynamics, the entire computation graph $G$ crossing multiple steps is fully differentiable.


In contrast to BC that minimizes the action distance $D_a(a \sim \pi_\theta || a^* \sim \pi_\textrm{exp})$ between the learner and the expert, we are minimizing the state distance $D_s(s \sim \tau_\theta || g \sim \tau_\textrm{exp})$ between the learner agent trajectory and the expert trajectory by differentiating through the temporal computation graph $G$. Such a design choice enables \aname with self-supervision in the unseen environment and hence addresses the covariant shift issue. To minimize the distance between the learner agent trajectories and the expert demonstrations, the first option is to apply a direct L2 loss between them, and back-propagate the gradient through the differentiable dynamics: 
\begin{equation*}
    \textrm{arg} \min_{\theta} 
    \sum_{s \sim \tau_\theta} 
    \sum_{t=0}^T (g_t - s_t)^2.
\end{equation*}
Based on the L2 loss objective function,  each state $g_t$ in expert demonstrations can be considered as a local learning goal at the time step $t$ for the leaner agent to achieve. 
However, the objective function above suffers an issue of enforcing exact match between the state $s_t$ in leaner policy trajectories and the state $g_t$.
The corresponding learning goal $g_t$ for each $s_t$ may be impractical to achieve as the $s_t$ and $g_t$ can be far away at the beginning of the training state. 
However, such impractical goals exceed the capability that the differentiable dynamics can offer and hence often result in local optima.

\algdef{SE}[SUBALG]{Indent}{EndIndent}{}{\algorithmicend\ }%
\algtext*{Indent}
\algtext*{EndIndent}
\begin{algorithm}
    \caption{\methodname }
    \begin{algorithmic}[1]
        \Require I \quad Optimization Iteration  \par
        \Ensure The best estimated policy
        
        \State Collect \(J = 1\) expert demonstrations.
        \State Initialize the stochastic policy as \(\pi_\theta\).
        \State Pretraining $\pi_\theta$ using Behaviour Cloning.
        \For{ optimization iteration $i = 1 \cdots I $}
        \Statex \quad  \quad \# Evaluate and optimize policy

        \State Roll out trajectories \(\tau_\theta\)
        \State Compute loss function L:
        
        \begin{equation*}
            L = \frac{1}{|\tau_\textrm{exp}|} \sum_{g_t \in \tau_\textrm{exp}} \min_{s \in \tau_\theta} \| g_t - s \|_2^2  
            +\alpha \frac{1}{|\tau_{\theta}|} \sum_{s_t \in \tau_{\theta}} 
            \min_{g \in \tau_\textrm{exp}} \| g - s_t \|_2^2
        \end{equation*}
        
        \State Update the policy \(\pi_\theta\) with analytical gradient \(\bigtriangledown_\theta L\).

        \EndFor
        \State \textbf{Return} the policy \(\pi_\theta\).
    \end{algorithmic}
        \label{alg:DIL}
\end{algorithm}

\subsection{Imitation Learning via Differentiable Physics}




Considering the impractical learning goals in the L2 objective, we develop a new approach called \methodname (\aname). \aname considers the states in the imitation learning task as a set of unordered points and matches them to the expert demonstration. 
We introduce Chamfer-$\alpha$ loss for trajectory matching with the expert. Instead of selecting those faraway correct but impractical goals, \aname dynamically selects the nearest local goals to the demonstrated states,
which gives a simpler optimization landscape. Specifically, Chamfer-$\alpha$ loss can be separated into two parts, deviation loss and coverage loss.


\textbf{Deviation loss}. The Equation \eqref{eq:local_goals} shows the deviation loss function $L_\textrm{d}$ for selecting the suitable goals. For each state $s \in \tau_\theta$, we treat it as a local optimization problem and set the learning goal to the nearest state $g \in \tau_\textrm{exp}$. The intention is to select the easier goals for the agent to follow. Concretely, this helps to constrain the gradient scale such that we can obtain a more stable optimization process during the BPTT with physics operators.
The summation term measures the deviation loss between the learner policy roll-outs and expert demonstration. The intuition is that the learned policy should produce states similar to those from the expert policy and not deviate too much. We present the deviation loss as follows:
\begin{equation}
    L_\textrm{d} = \frac{1}{|\tau_{\theta}|} \sum_{s_t \in \tau_{\theta}} \min_{g \in \tau_\textrm{exp}} \| g - s_t \|_2^2.
    \label{eq:local_goals}
\end{equation}
However, deviation loss alone may cause the ``state collapse'' issue.
For example, it is possible that all $s_t \in \tau_{\theta}$ are close to a small subset of $\tau_\textrm{exp}$ with low deviation cost. As a result, deviation loss provides no coverage to the expert trajectory and is hence sub-optimal. Therefore, the ultimate goal is to learn $\pi_\theta$ that covers all the states of a trajectory $\tau_\textrm{exp}$ and at the same time stays close to the expert states. 



\textbf{Coverage loss}. 
To ensure all states in the expert trajectory are covered by the learner policy, we introduce an extra coverage loss in Eqn.~\ref{eq:coverage}:
\begin{equation}
    L_\textrm{g} =  \frac{1}{|\tau_\textrm{exp}|} \sum_{g_t \in \tau_\textrm{exp}} \min_{s \in \tau_\pi} \| g_t - s \|_2^2. 
    \label{eq:coverage}
\end{equation}
Intuitively, the coverage loss guarantees that each state in the expert trajectory be close to the rollout trajectory generated by the learner policy. This naturally alleviates the ``state collapse'' issue introduced by the deviation loss.
Combining the deviation loss and coverage loss, we introduce Chamfer-$\alpha$ loss function for \aname:
\begin{equation}
    L_\textrm{Chf-$\alpha$} = L_d + \alpha L_g = \frac{1}{|\tau_\textrm{exp}|} \sum_{g_t \in \tau_\textrm{exp}} \min_{s \in \tau_\theta} \| g_t - s \|_2^2  
    +\alpha \frac{1}{|\tau_{\theta}|} \sum_{s_t \in \tau_{\theta}} 
    \min_{g \in \tau_\textrm{exp}} \| g - s_t \|_2^2.
    \label{eq:objective}
\end{equation}
Although derived from different objectives, Eqn.~\ref{eq:objective} resembles the Chamfer distance for measuring the distance between two sets, which is widely used in computer vision~\cite{butt1998optimum,ma2010boosting,ChenS-RSS-21}. Thus, we name our loss as Chamfer-$\alpha$, which balances the ratio between deviation and coverage loss with a hyper-parameter $\alpha$.
In our experiments, we observe that having a lower $\alpha$ can shape the learned policy to imitate the expert at a global level and hence has a faster convergence. However, there is a trade-off between the convergence speed and final performance as the larger deviation factor converges slower but the final performance can be better. A visual illustration of Chamfer-$\alpha$ loss is shown in \figref{fig:loss_demo}.

We summarize our method in Algorithm \ref{alg:DIL}. We first collect $J=1$ episode expert demonstration and use the standard supervised behavior cloning to bootstrap the policy learning using the expert demonstration. Next, we roll out the dynamics with a large batch of learner policy trajectories in parallel to speed up the training. With the objective function \eqref{eq:objective} defined above, we compute the loss and update the policy parameters $\theta$.

\section{Experiment}

We evaluate our method on Brax~\cite{freeman2021brax} environments, providing a variety of differentiable MuJoCo-like continuous control tasks. In addition, we have developed a new robotic deformable object manipulation task, which requires hanging a piece of cloth on a stand, to demonstrate the generality of our approach to changing environment dynamics. 

We compare our approach with two state-of-the-art IL methods, PWIL \cite{dadashi2020primal} from inverse reinforcement learning (IRL) and DAC \cite{kostrikov2018discriminator} from adversarial imitation learning (AIL). We follow their official published implementation and change the evaluation tasks to differentiable environments. For a fair comparison, we change their subsampling rate to use a single expert episode for training.

In our experiments, we aim to answer the following questions: 1) Can our method \aname recover expert behavior? 2) How fast does \aname converge? 3) Is \aname generalizable to complex deformable object manipulation tasks with changing dynamics? 4) What are the core parameters that influence performance?

\begin{figure}[t]
	\centering
	\begin{tabular}{c c c c}
    \includegraphics[width=0.23\linewidth]{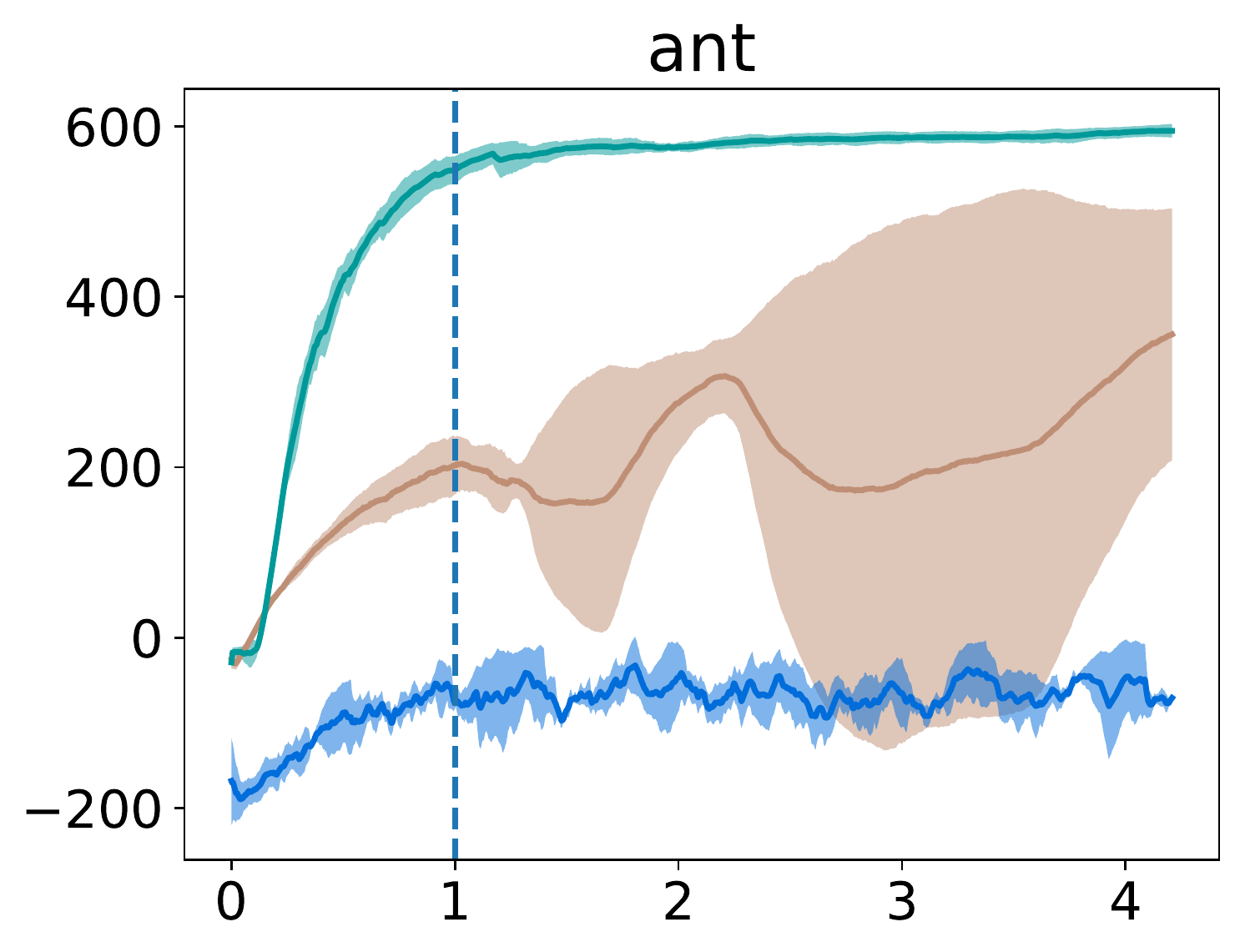} &
    \includegraphics[width=0.23\linewidth]{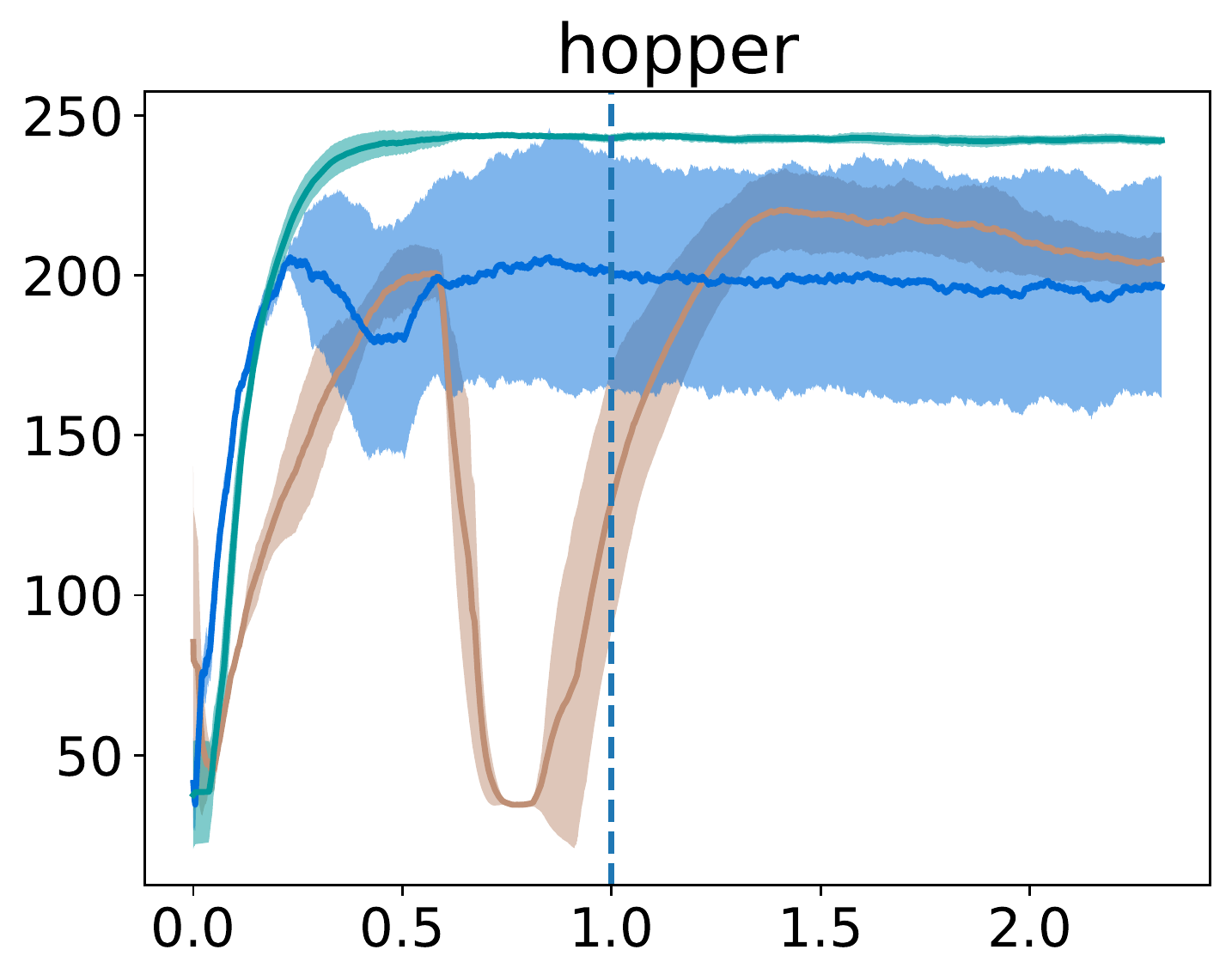} &
    \includegraphics[width=0.23\linewidth]{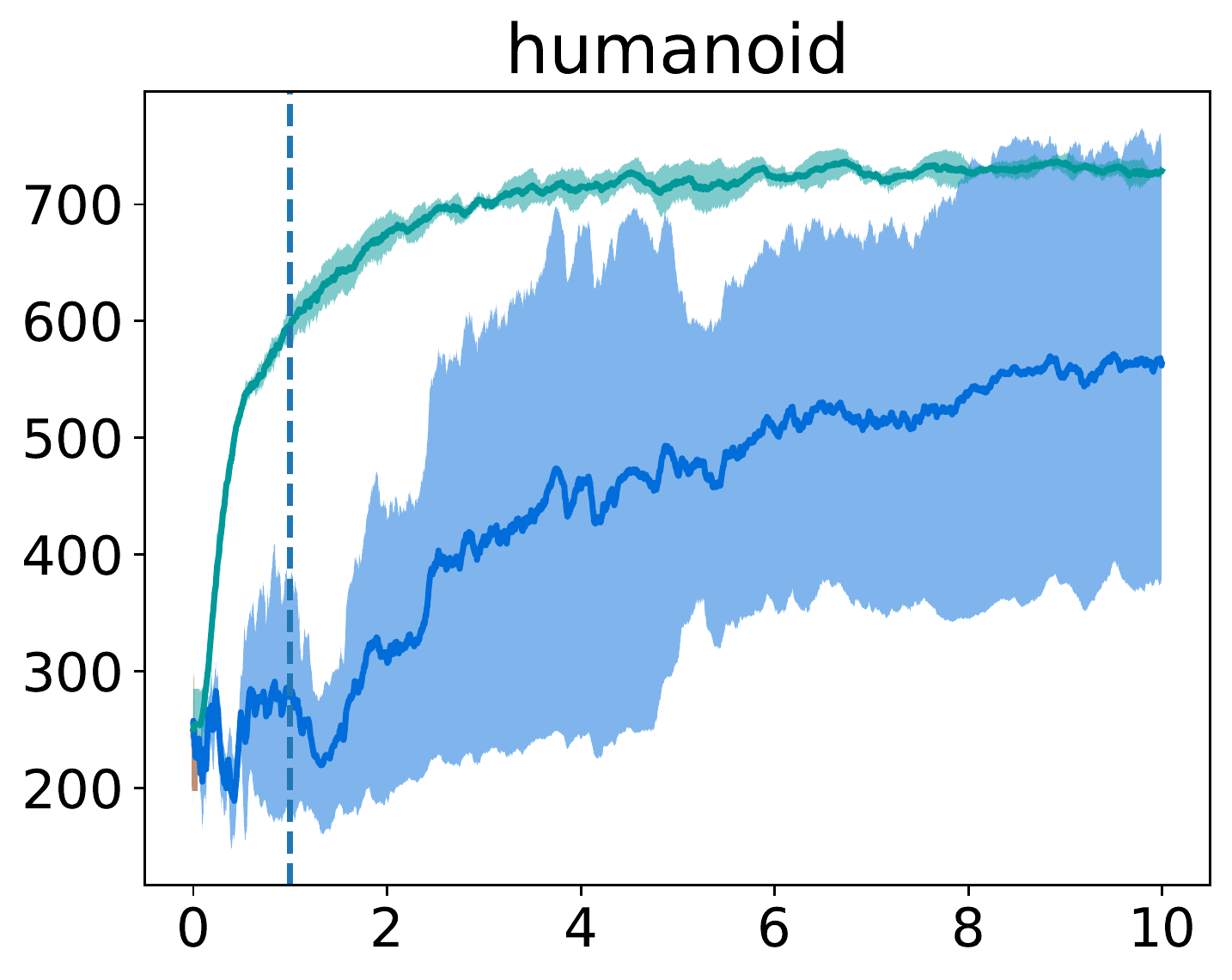} &
    \includegraphics[width=0.23\linewidth]{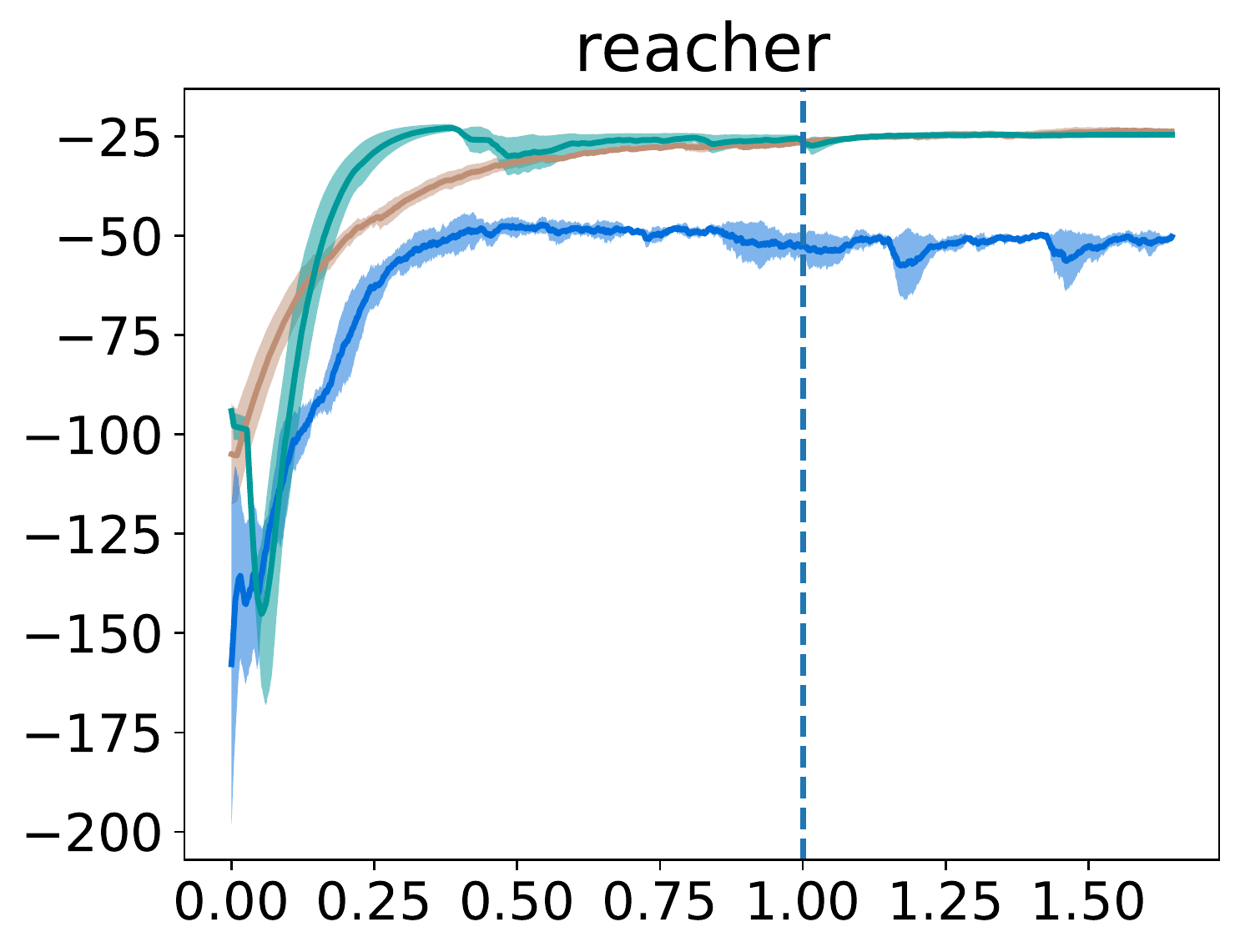} \\
    
    \includegraphics[width=0.23\linewidth]{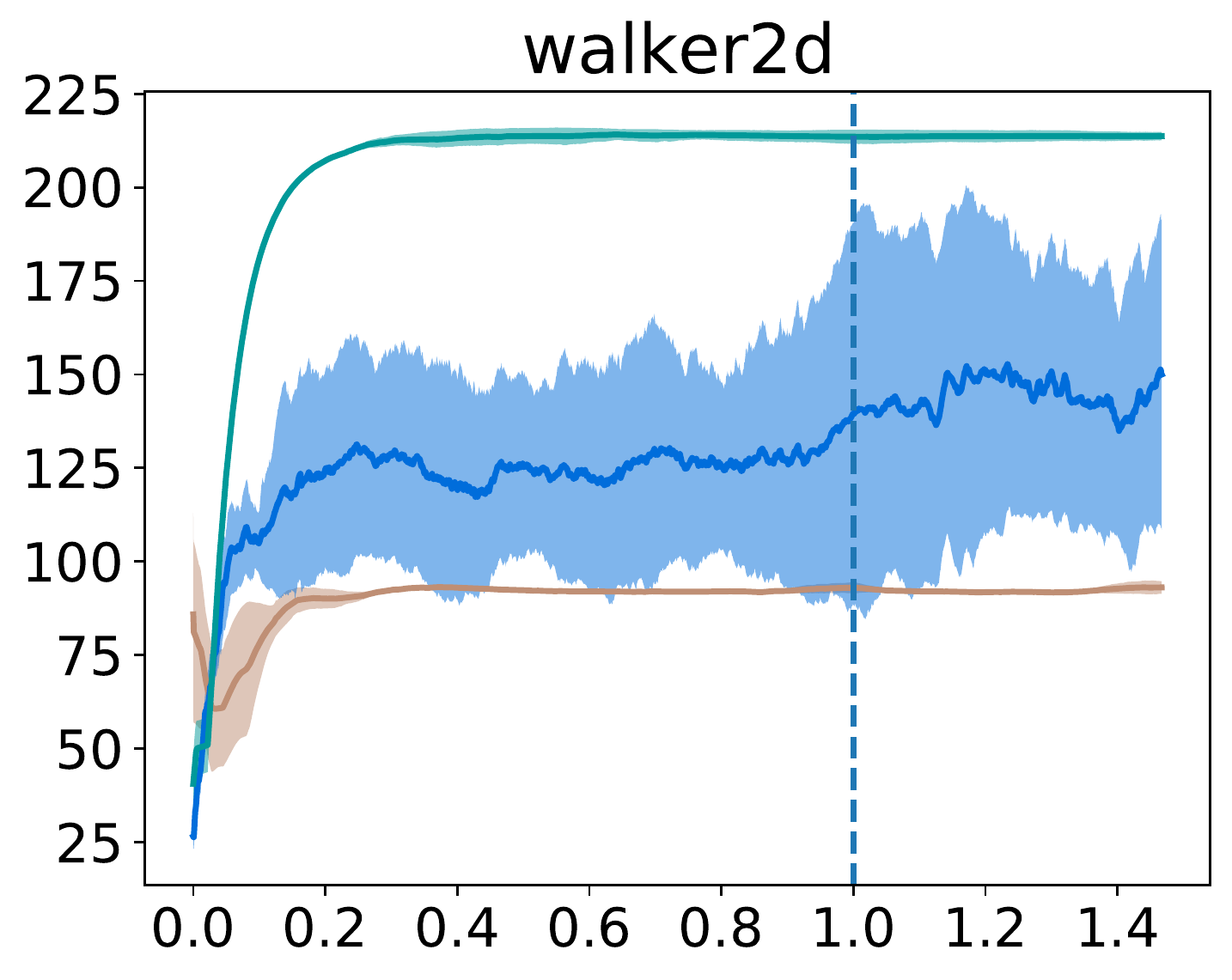} &
    \includegraphics[width=0.23\linewidth]{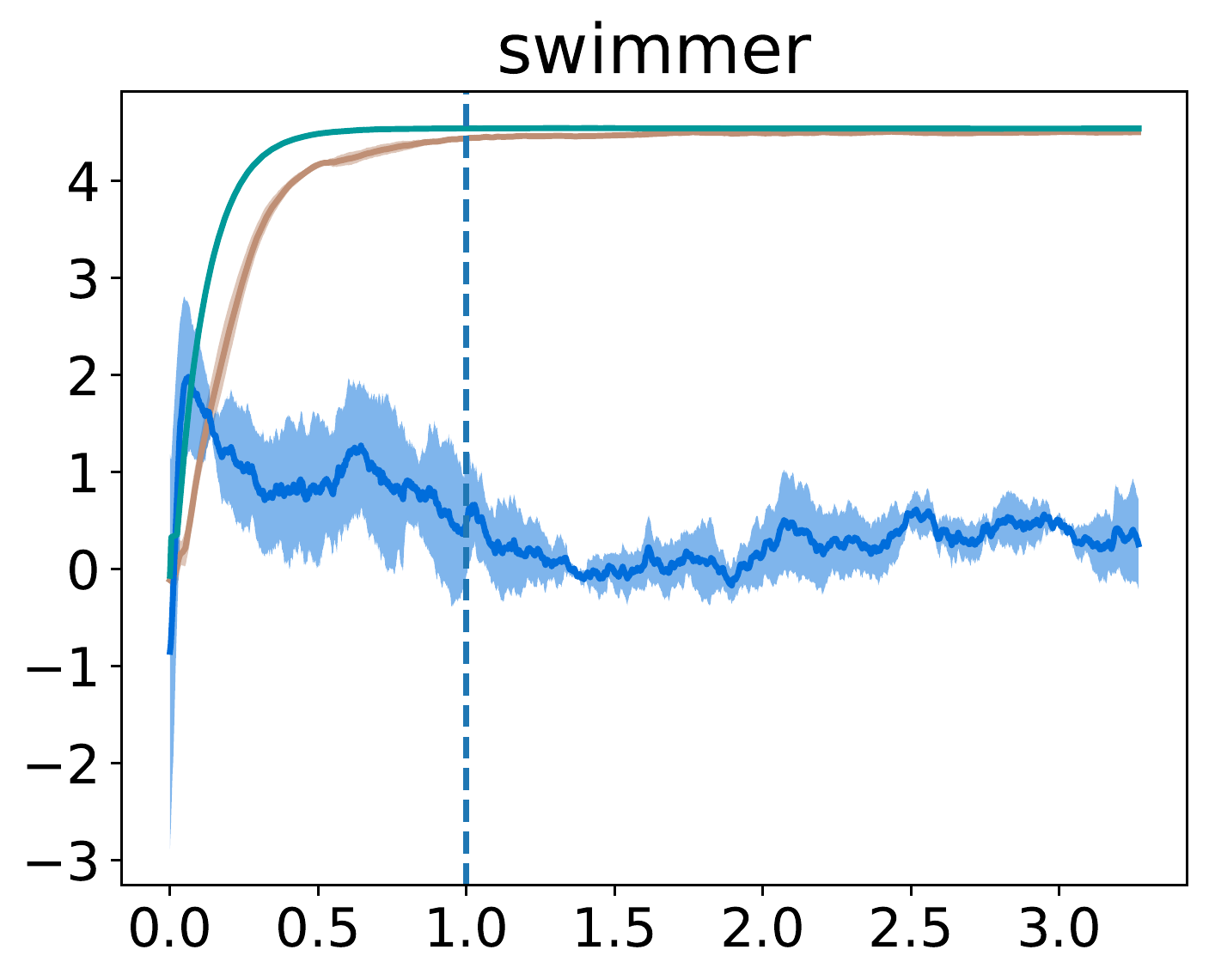} &
    \includegraphics[width=0.23\linewidth]{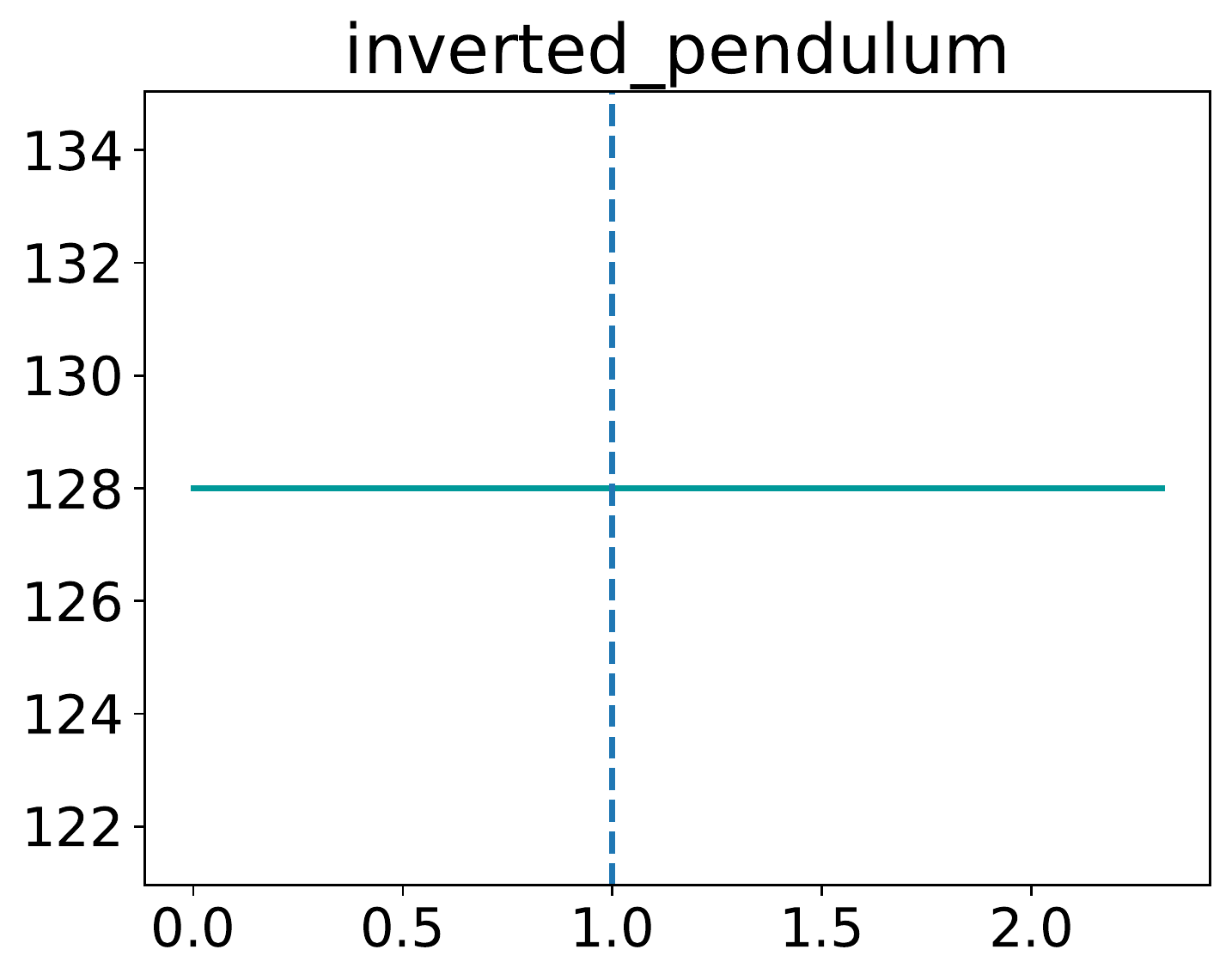} &
    \includegraphics[width=0.23\linewidth]{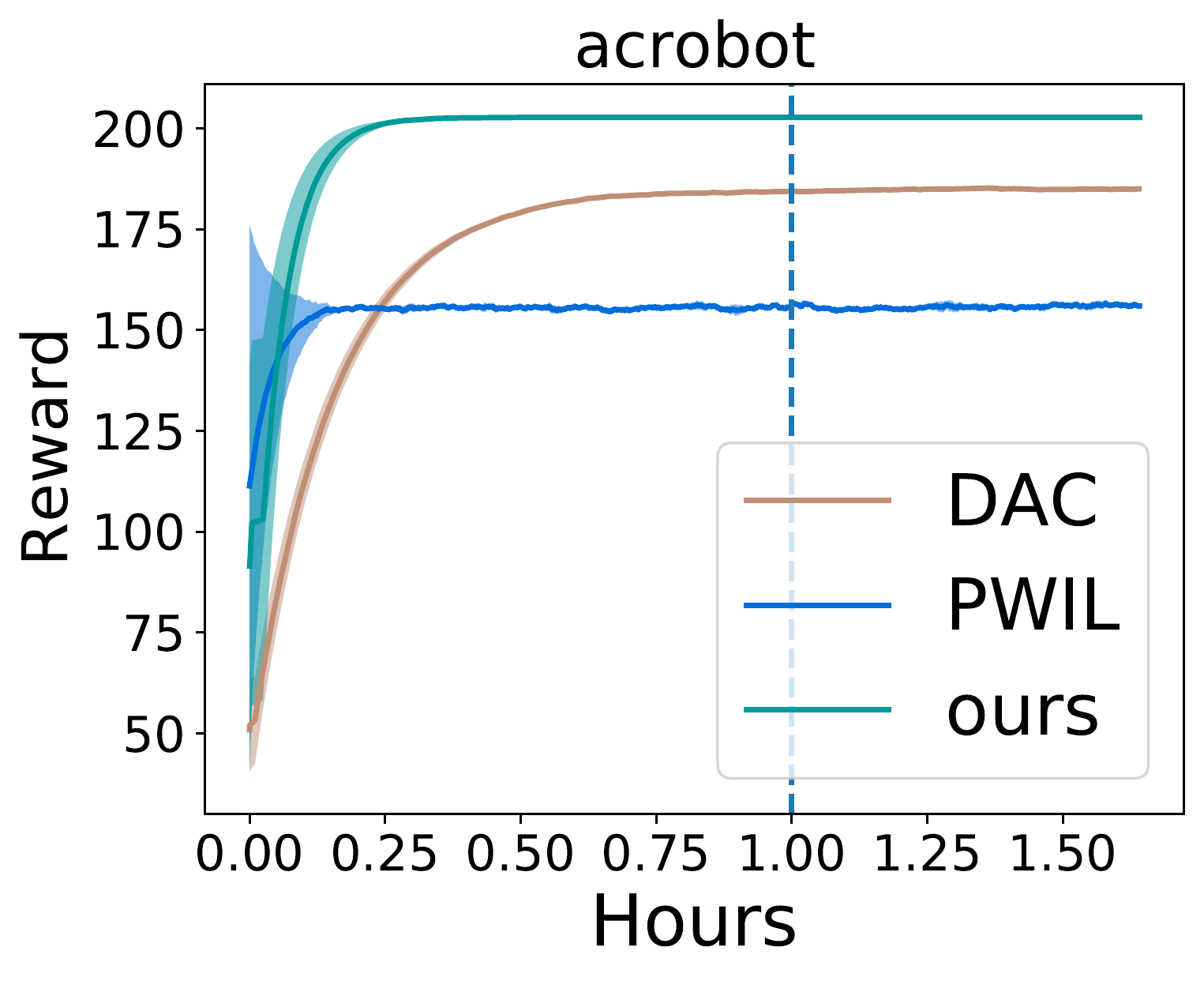} \\
    
	\end{tabular}
	\centering
	\caption{\small Relative wall time performance. We evaluate \aname on the Brax MuJoCo \cite{freeman2021brax} tasks and only one expert demonstration is provided to all the IL methods. 
    The results show that our method generally outperforms the SOTA methods, with a much faster convergence speed, a more stable learning process, and smaller variance.
	The differentiable dynamics gives a single-loop training process and avoids the noisy intermediate signal. In addition, in difficult tasks such as ant, walker2d and humanoid, \aname outperforms the other baselines by a large margin given limited training time, because the differentiable physics helps to reserve more information embedded in the trajectories.
	Given an hour of training time (indicated by the vertical dashed line), we achieve 36\% higher performance on average over all the tasks and baselines.}
	\label{fig:wall_time}
\end{figure}

\subsection{Brax Continuous Control Tasks}

We evaluate our approach on the 8 Brax continuous control MuJoCo-like tasks: Ant, Hopper, Humanoid, Reacher, Walker2d, Swimmer, Inverted Pendulum, and Acrobot. For each task, we train a PPO \cite{schulman2017proximal} agent to act as an expert. The episode length for all tasks is 128, and we collect only one expert episode for each task, which is used to train the IL agent. Each number reported in the result table is evaluated with 3 different seeds.

\textbf{Implementation Details}. 
In contrast to the IRL and AIL methods, our method \aname has only one policy network consisting of three MLP layers with Swish activation. 
The number of their hidden neurons is 512, 256, and the corresponding action dimension of the task, respectively.
We clip the gradients with a maximum gradient norm value of 0.3 to regularize the learning process. To speed up the convergence, we use a batch size of 360 on an NVIDIA A100 graphics card. The deviation factor $\alpha$ and gradient truncation length are set to 1 and 10, respectively. We train our policy network with an Adam optimizer with a learning rate of 0.001 for 5,000 updates. The entire script is written by Flax~\cite{flax2020github} and JAX~\cite{jax2018github}. For a fair comparison, all methods use the same amount of computational resource.

\textbf{Results and Discussions}.
The overall performance is presented in \tabref{table:brax_results} and the detailed training curves are given in \figref{fig:wall_time}. We observe that \aname outperforms the SOTA methods on 6 out of 8 tasks, and achieves comparable performance on the remaining 2 tasks. Also, with only one expert demonstration, \aname generally recovers the expert behavior and mostly achieves comparable performance. Specifically, we notice that the performance gain of \aname increases as the complexity of the task increases. For example, \aname achieves comparable performance with the baselines on the relatively easy task of Inverted Pendulum but outperforms 
the baselines on Ant environment. In addition, \figref{fig:wall_time} shows that \aname has much lower variance and a more stable training curve compared with the SOTA methods. We believe both of the above advantages come from the benefits of differentiable physics. By back-propagating the gradients from states directly to the policy, \aname avoids the dynamic target introduced by the intermediate signals and stabilizes the training process; by considering the physics prior for policy learning, \aname obtains a policy that generalizes better to complex dynamics.


\begin{table}[t]
\centering
\caption{\small Brax MuJoCo Task Results}
\fontsize{8}{8}\selectfont
\begin{tabular}{ccccccccc}
\toprule

& Ant & Hopper & Humanoid & Reacher & Walker2d & Swimmer & Inverted pendulum & Acrobot \\

\midrule
DAC & 393.57 & 220.54 & 256.63 & \textbf{-21.52} & 93.19 & 4.52 & \textbf{128.00} & 185.26  \\ 
PWIL & -1.98 & 205.72 & 722.41 & -47.19 & 205.78 & 1.98 & \textbf{128.00} & 157.03  \\ 
ILD(ours) & \textbf{594.88} & \textbf{243.93} & \textbf{736.87} & -22.86 & \textbf{214.17} & \textbf{4.54} & \textbf{128.00} & \textbf{202.74}  \\ 
\midrule
Expert & 624.34 & 292.83 & 933.24 & -22.49 & 289.14 & 4.29 & 128.00 & 200.80 \\

\bottomrule
\end{tabular}
\label{table:brax_results}
\end{table}

Moreover, although AIL and IRL methods benefit from the sample efficiency of the small number of interactions with the environment, they often suffer long training time due to the double-loop structure. Such a double-loop structure involves additional computations and slows down the training process. In contrast, \aname has a single-loop design and allows the policy to be optimized directly with the differentiable dynamics. In \figref{fig:wall_time}, we observe that \aname converges significantly faster than both DAC and PWIL with a much smoother and stable training curve.



\subsection{Robot Cloth Manipulation}

\begin{figure}[t]
	\centering
	\begin{tabular}{c c}
    \includegraphics[width=0.45\linewidth]{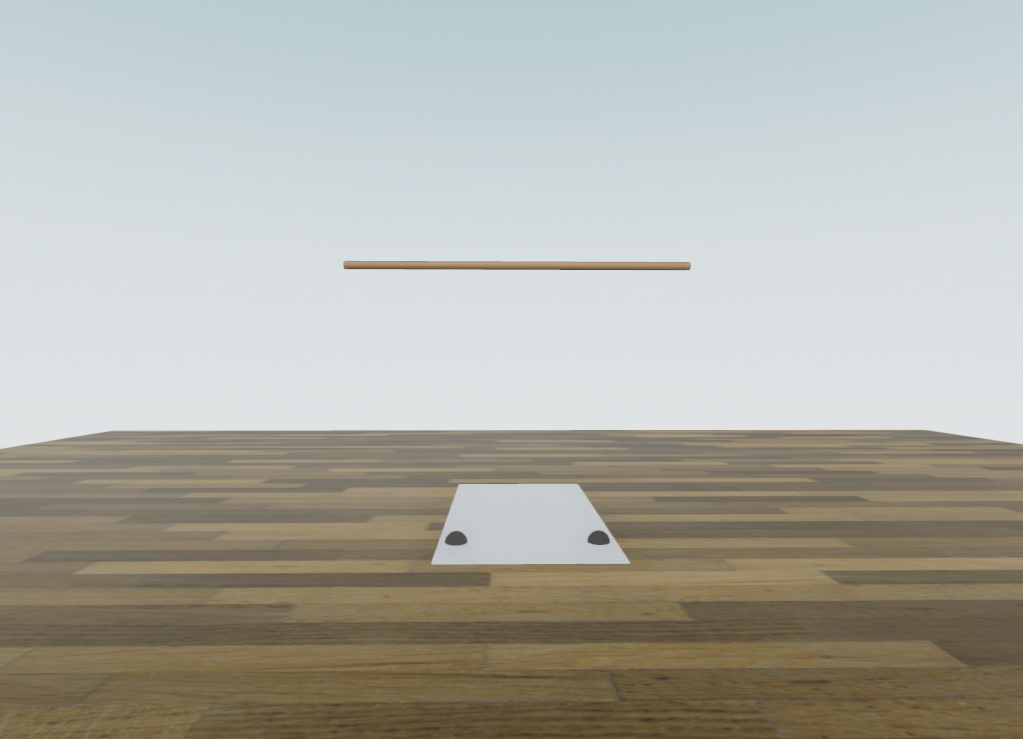} &
    \includegraphics[width=0.45\linewidth]{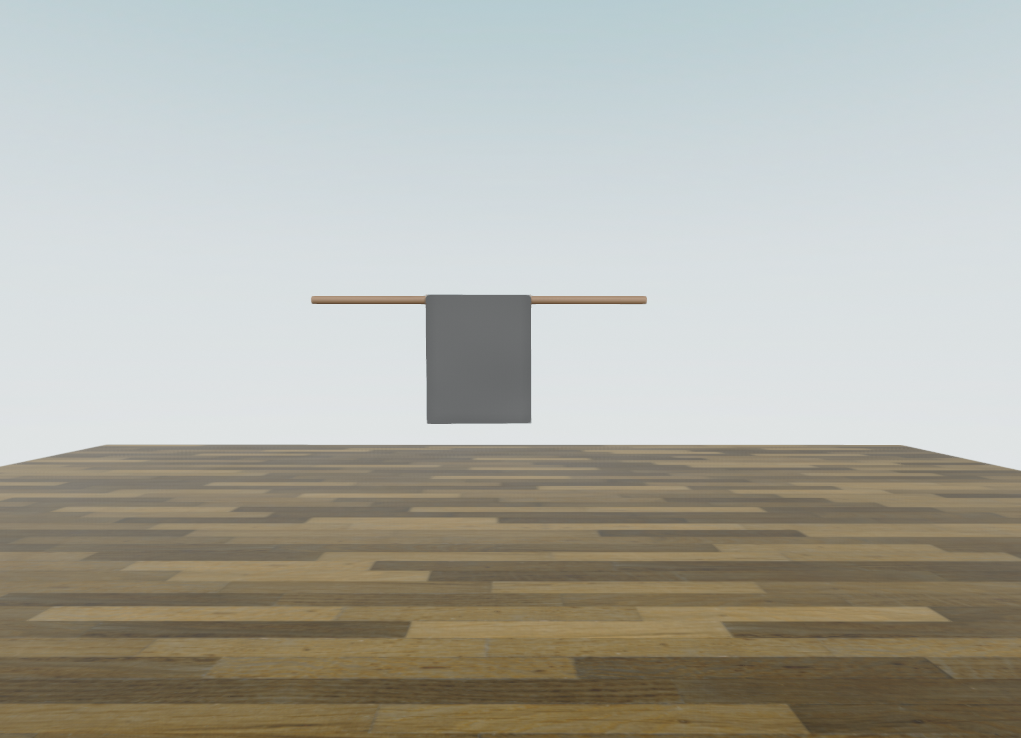} \\

    (a) \small Initial State & (b) \small Goal State
    
	\end{tabular}
	\centering
	\caption{\small Deformable cloth manipulation. The task is to control two grippers (dark color) to hang the piece of cloth on the pole. Image (a) shows the initial state and image (b) shows the target state. This task is challenging because the dynamics are complex and the observation space is huge, with 1,736 dimensions.}
	\label{fig:cloth}
\end{figure}

In this section, we test whether our approach can be generalized to a more complex robotic deformable object manipulation task, where a piece of cloth is to be hung on a pole. A visualization of this task can be found in \figref{fig:cloth}. The main challenge of this deformable object manipulation task is the changing dynamics, where the testing dynamics are different from the demonstrating dynamics.  Specifically, we add additional bias and noise to the input actions of the test environment, making the dynamics of the test environment different from the dynamics demonstrated by experts. As a result, we obtain an environment with noisy dynamics and fixed goals. 
In addition, the high-dimensional state space of deformable objects makes it hard to extract useful features for policy learning. 
To collect demonstrations, we use a handcrafted policy as the expert policy. Similarly, only one trajectory is used as the demonstration.

\begin{figure}[h]
	\centering

    \includegraphics[width=0.4\linewidth]{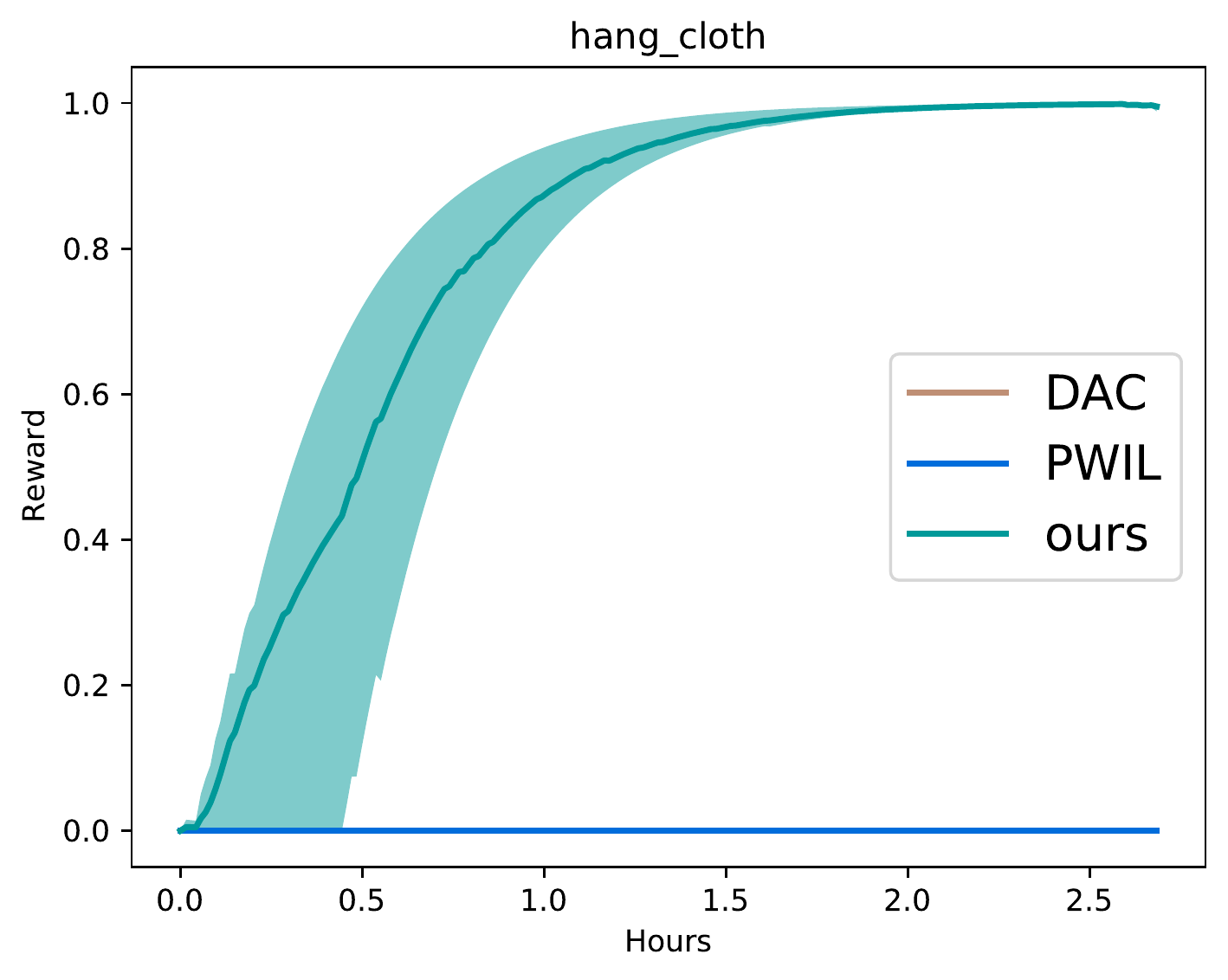}
    
    
	\caption{
	\small Deformable cloth manipulation Results. The task of manipulating the cloth has a very sparse reward: if the cloth is on the pole at the last step, the reward is 1; otherwise, all steps are 0. Therefore, the average reward can be interpreted as the task success rate. Compared to the expert demonstration environment, additional noise is added to the test environment, so the dynamics of the test environment change. Despite the dynamic change, our method \aname converges quickly and adapts to the new environment, while the other two baselines cannot be transferred in the dynamic change. }
	\label{fig:cloth_results}
\end{figure}

\textbf{Implementation Details}. We develop a cloth dynamics engine in JAX following the implementation of Taichi~\cite{hu2019taichi}.
The observation space for this task has 1,736 dimensions and consists of 288 key nodes on the cloth and 2 gripper states. The action space consists of 6 dimensions that control the speed of the two grasps. We assume that the two grippers have grabbed the two corners of the cloth.  
To facilitate the evaluation, we define a reward function that is 1 if the cloth is on the pole at the last step and 0 otherwise. This reward function also indicates the success rate of the training agent. The episode length for this task is 80 and a single expert demonstration is provided for all methods. We use the same implementation as the Brax environment with 3 MLP layers. In the complex task setting, we reduce the batch size from 360 to 50 due to hardware memory limitations. The learning rate is set to $1e^{-4}$ and the rest is the same. More details of the cloth simulation can be found in Appendix.

\textbf{Results and Discussions}. The results of the cloth manipulation can be found in \figref{fig:cloth_results}. The results show that our method can learn a stable policy in changing dynamics under complex observation conditions. We outperform the other two baselines, completing the task with a success rate of 1. Specifically, \aname achieves fairly stable performance with low variance even if the environment dynamics are constantly changing. The success of \aname comes from the physics prior implicitly encoded in the learner policy, which learns a distribution of the environment specifications and can recover the expert behavior even with unseen configurations.
The other two baselines fail in the task, even though we have tried hard to tune the parameters such as reducing their learning rate. It is probably because they have learned a dynamics-coupled reward function or discriminator. When the dynamics change, the reward function/discriminator they have learned cannot be transferred, thus causing the failure. In contrast, our approach is not coupled with specific dynamics and thus has better generalization. 

With the recent advances and growing interests in differentiable robot simulators, such as PlasticineLab~\cite{huang2021plasticinelab}, DiSECt~\cite{heiden2021disect}, and Scalable Diff~\cite{Qiao2020Scalable}, we believe our method could be a good starting point for exploiting the differentiability of physics simulators to boost the development of robot learning algorithms.





\subsection{Ablation Study}

In this section, we discuss the core hyperparameters and algorithm choices that may influence performance. The results can be found in \tabref{table:ablation_results}. 
The reported values in \tabref{table:brax_results} are trained using the Chamfer-$\alpha$ loss with a truncation length of 10, batch size of 360, and deviation factor of 0.2.

 \textbf{Loss Function}. We first compare the loss functions. We replace the original Chamfer-$\alpha$ loss with a simple step-wise L2 loss to compute the distance between the expert trajectory and the rollout trajectory. The results show that the performance of L2 loss is significantly lower compared to our method. This is because L2 loss simply performs a step-wise matching and ignores the actual distance between the states. As a result, this often introduces goals that are faraway from the current states and increases the optimization difficulty, especially through the BPTT over complex physics operators. Thus, it leads to suboptimal behaviors, which is a consistent observation with Freeman et al.~\cite{freeman2021brax}. In contrast, the proposed Chamfer-$\alpha$
 distance selects the local goal that matches each state, thus reducing the complexity of learning and giving a smoother optimization landscape.

\textbf{Truncation Length}. We test with different gradient truncation lengths. As shown in the ablation results table, we find that short truncation lengths such as 1 usually lead to poor performance because in this case, the policy is trained by one-step signal and ignores the multi-step dynamics of the environment. However, large truncation lengths tend to lead to gradient explosion and in the end hinder the stability of the learning process. Nevertheless, when developing the cloth simulation environment, we observe that if with step-wise gradient normalization, the gradient explosion issue can be significantly alleviated and the performance of \aname can be generally improved. Empirically, a truncation length of 10 is a safe choice and often brings good performance.

\textbf{Deviation Factor}. We evaluate the influence of different deviation factors, which balance the deviation loss and coverage loss. We observe that the best performance is achieved with a deviation factor 1. However, we also observe that the smaller the factor is, the faster the global convergence will be, which introduces the trade-off between the final performance and the convergence speed. In our reported values, we use a deviation factor of 1, but the deviation factor 0.2 produces a comparable performance with a faster convergence speed, which could be considered for time sensitive applications. More details are available in the appendix.



\begin{table}[t]
\centering
\caption{\small Brax MuJoCo Ablation Results}
\fontsize{7}{7}\selectfont

\begin{tabular}{ccccccccccc}
\toprule


& \multicolumn{2}{c}{Loss} 
& \multicolumn{4}{c}{Trunc Length}
& \multicolumn{4}{c}{Deviation Factor}\\
\cmidrule(lr){2-3}\cmidrule(lr){4-7}\cmidrule(lr){8-11}
& Chamfer-$\alpha$  & L2    & 1 & 10 & 30 & 100     & 0 & 0.2 & 1 & 5\\

\midrule

ant & \textbf{583.77} & 514.07 & 110.07 & \textbf{583.77} & -15.50 & -16.29 & 560.95 & 583.77 & \textbf{594.88} & 552.25  \\ 
hopper & \textbf{242.27} & 173.39 & 53.45 & \textbf{242.27} & 217.87 & 144.95 & 239.96 & 242.27 & 243.93 & \textbf{248.76}  \\ 
humanoid & \textbf{715.14} & 542.94 & 331.27 & 715.14 & \textbf{788.84} & 355.09 & 704.96 & 715.14 & \textbf{736.87} & 710.12  \\ 
reacher & -23.18 & \textbf{-22.02} & -31.72 & -23.18 & -23.24 & \textbf{-21.99} & -75.36 & -23.18 & \textbf{-22.86} & -22.95  \\

\bottomrule
\end{tabular}
\label{table:ablation_results}
\end{table}

\subsection{Limitations}
Our approach relies on differentiable dynamics, but not all tasks are naturally differentiable. 
Many games, such as Atari games and board games, are non-differentiable and therefore our approach does not apply.
However, in the real world there are many valuable tasks that are indeed differentiable, such as robotic deformable object manipulation. In addition, a lot of new differentiable physics engines have emerged, such as PlasticineLab~\cite{huang2021plasticinelab} for soft deformable objects, DiSECt~\cite{heiden2021disect} for knife cutting, and Scalable Diff~\cite{ Qiao2020Scalable} for cloth and soft objects manipulation. 
Another limitation is the gradient truncation length, where too short truncation length does not perform well, and too large can result in gradient explosion. However, if the differentiable dynamics normalizes its gradient in every step-wise transition function call, the gradient exploding issue can be significantly alleviated. 

\section{Conclusion}

In this work, we identify the benefits of differentiable dynamics and propose to use differentiable dynamics to learn IL agents. The core advantage of our approach is to move away from the traditional double-loop learning design and avoid the noisy intermediate learning signal.
The differentiable dynamics provides us with a new type of IL algorithm and brings better performance and remarkably lower variance of the learned IL agent. 
The use of Chamfer-$\alpha$ distance enables dynamic selection of local targets, significantly reducing the learning difficulty and giving better performance. In future work, we will further address the sample efficiency issue by using small batches and short rollout trajectories. At the same time, we should target more challenging tasks, such as manipulation tasks of various robotic deformable objects. We conclude that our IL learning method has only a single learning loop, but outperforms other IL baselines. In addition, we demonstrate that our approach has great potential for more challenging but valuable robotic deformable manipulation tasks. 

\section*{Acknowledgements}
This research is supported by the National Research Foundation, Singapore under its AI Singapore Programme (AISG Award No: AISG2-PhD-2021-08-015T).

\bibliographystyle{abbrv}
\bibliography{neurips_2022}
\newpage

\section*{Appendix}

\section{Deviation Factor}
\label{sec:training_details}

\begin{figure}[h]
	\centering
	\begin{tabular}{c c c c}
    \includegraphics[width=0.22\linewidth]{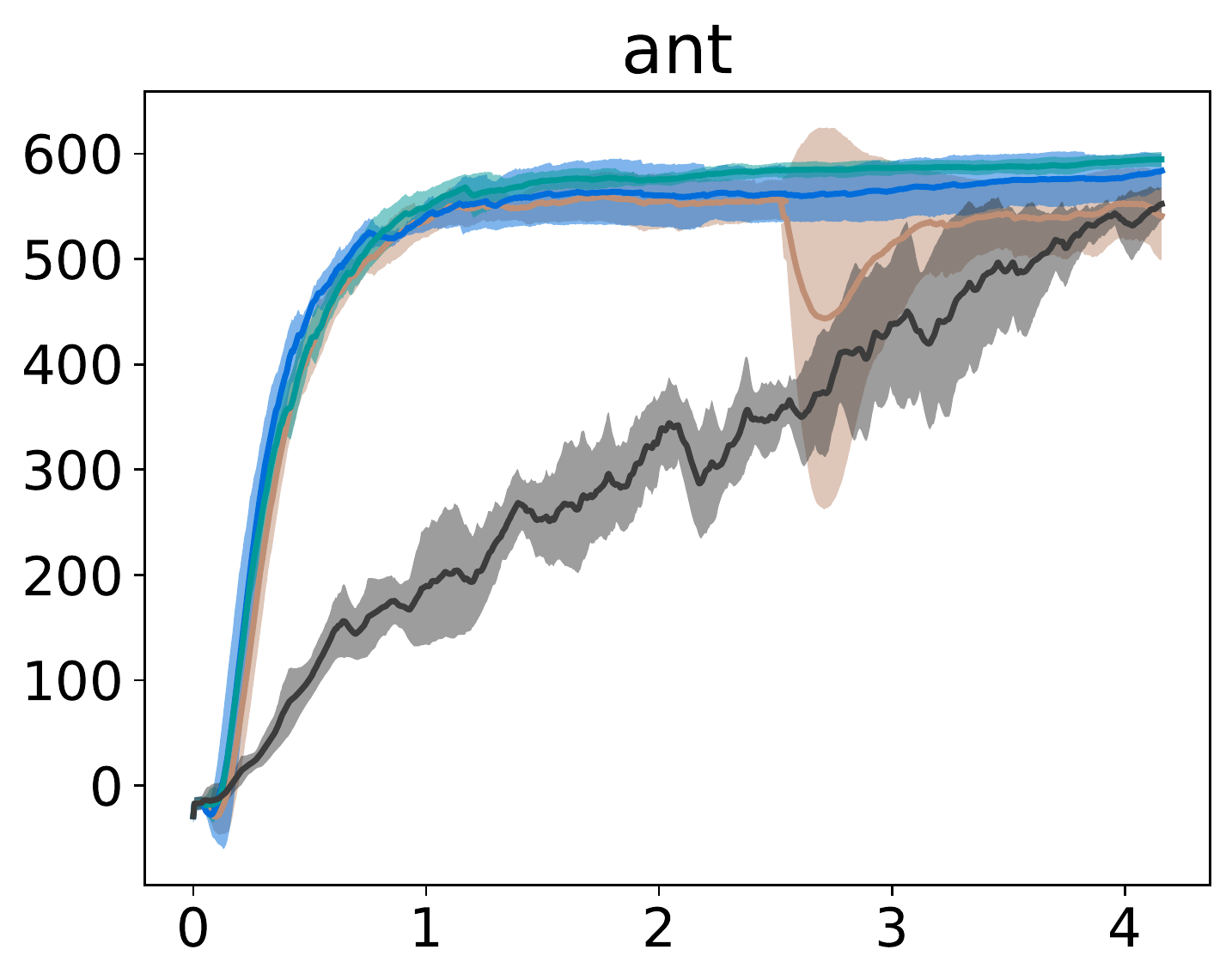} &
    \includegraphics[width=0.22\linewidth]{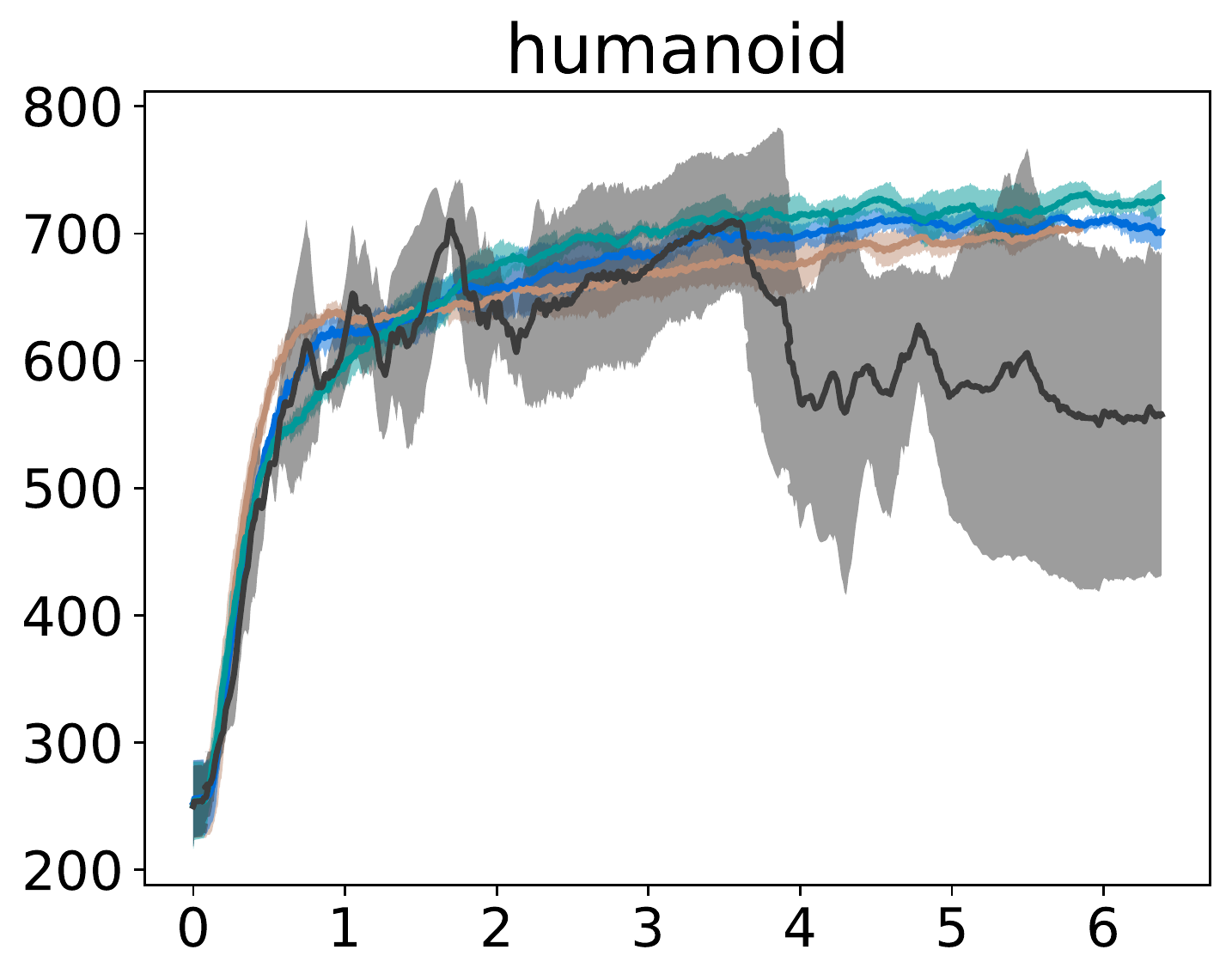} &
    \includegraphics[width=0.22\linewidth]{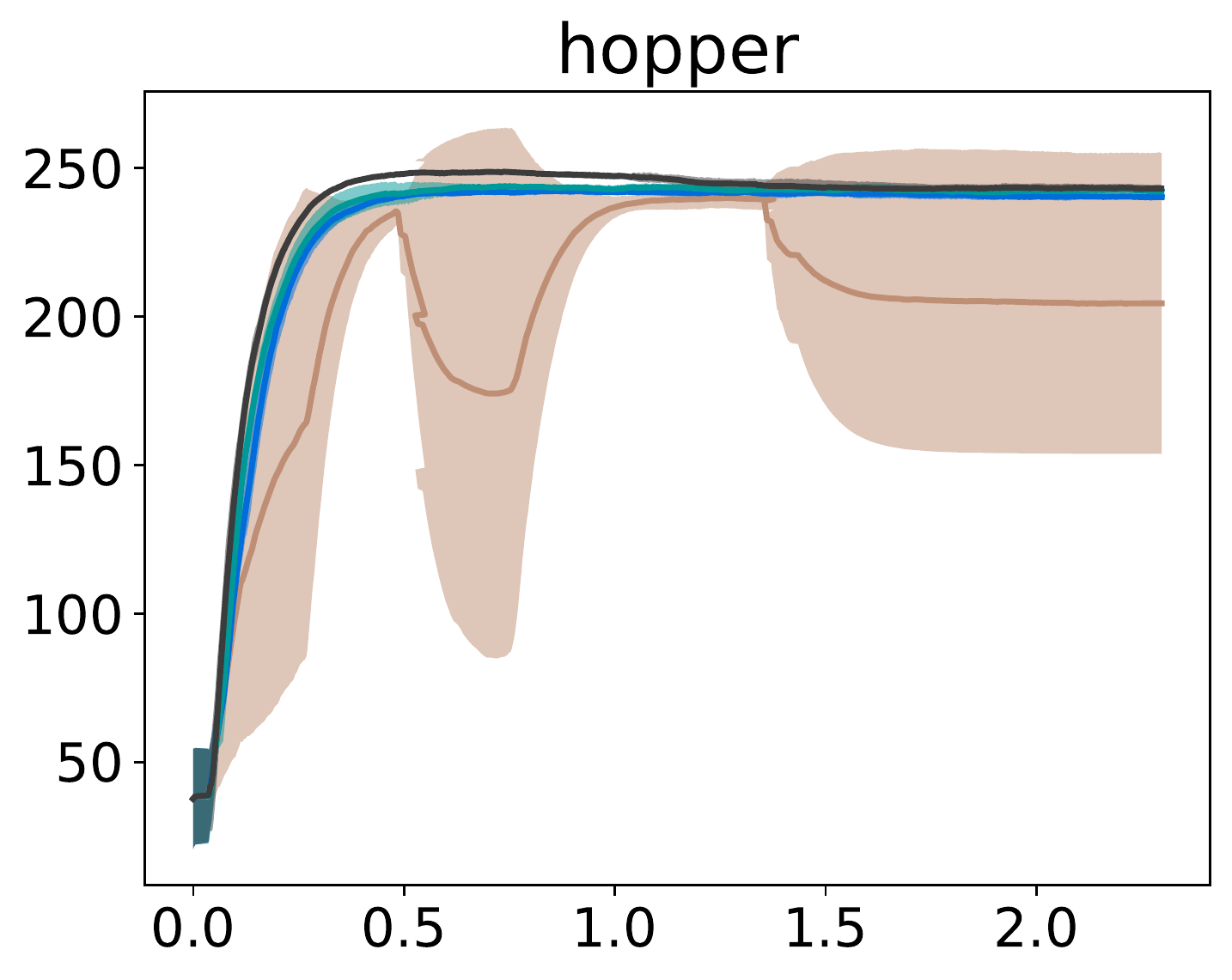} &
    \includegraphics[width=0.22\linewidth]{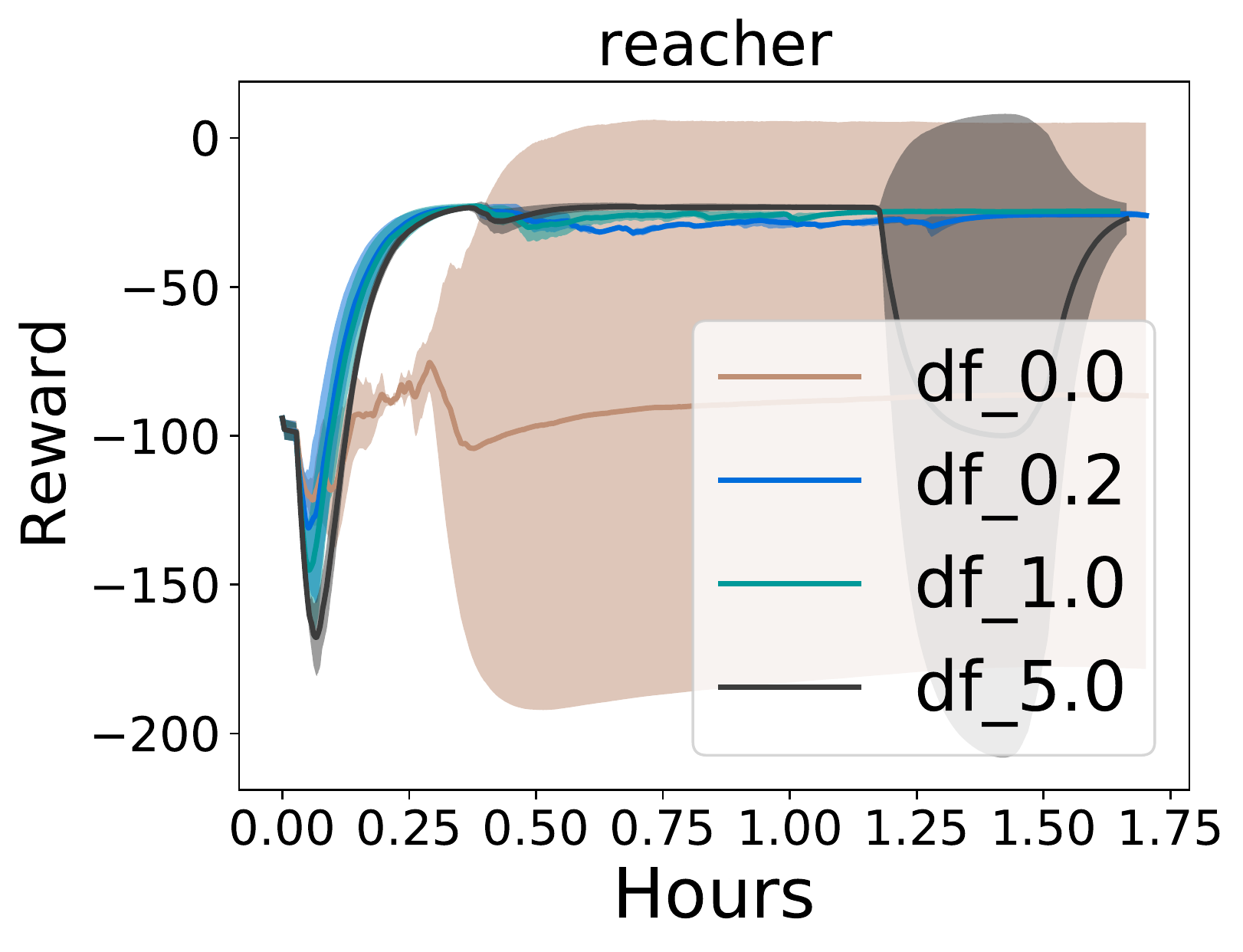} \\
    
	\end{tabular}
	\centering
	\caption{\small The training curves with different deviation factors. Setting the deviation factor to 0 or 5 will make the training unstable. However, \aname is less sensitive to moderate deviation values such as 0.2 to 1.   }
	\label{fig:df}
\end{figure}

In \figref{fig:df}, we show more details about the deviation factor. In general, if we do not force the learner state to be close to the expert state, i.e., set the deviation factor to 0, then the training process is unstable and tends to be suboptimal. On the other hand, if we focus too much on local state matching and set the deviation factor to 5, the learner policy tends to be conservative and unstable. Between these two, \aname is robust to deviation factors from 0.2 to 1 and does not vary much. In addition to this, if we compare 0.2 with 1.0, the smaller deviation factor learns slightly faster than the larger value, however, as a trade-off, its final performance is lower.

\section{Cloth Simulation}
\label{sec:cloth_sim}

Our cloth simulator is written in Jax and developed on top of the Taichi~\cite{hu2019taichi} implementation. As shown in \figref{fig:cloth}, a piece of cloth is lying on the ground and the goal is to put this cloth on a pole by controlling two black grippers. The state of the cloth consists of 288 key points in the shape of $(288,6)$, where the 6 dimensions are position and velocity. The underlying physics engine is built on Hooke's law, as shown below:
\begin{align*}
    f_i &= \sum_j -k (||x_i - x_j||_2 - l_{ij})(x_i - x_j)\\
    v_{t+1} &= v_t + \Delta t \cdot  \frac{f}{m}\\
    x_{t+1} &= x_t + \Delta t \cdot v_{t+1}
\end{align*}
where $f_i$ is the force at the $i_\textrm{th}$ point, $j$ refers to the $j_\textrm{th}$ neighbor, $x_i$ is the position of the $i_\textrm{th}$ point, and $l_{ij}$ is the rest distance. In general, the longer the stretching distance, the higher the resistance force. By averaging the forces of all neighbors, we can calculate the next state of the point. In more detail, we set $\Delta t$ to 2e-3 and repeat the above update equation 50 times for each robot action input. Thus, the dynamics of the deformable object is accumulated through time and the computational graph is long. To alleviate the gradient explosion and gradient vanishing problems, we normalize the gradients at each step of the backpropagation process.

\end{document}